\documentclass[twocolumn,pre,floatfix,superscriptaddress]{revtex4}
\usepackage[T1]{fontenc}
\usepackage[latin9]{inputenc}
\usepackage{mathrsfs}
\usepackage{amsmath}
\usepackage{amssymb}
\usepackage{graphicx}
\usepackage{braket}
\usepackage{color}
\usepackage[svgnames]{xcolor}
\usepackage{comment}
\usepackage{natbib}
\usepackage{dsfont}
\usepackage{mdframed}

\makeatletter

\usepackage{listings}

\usepackage{hyperref}

\hypersetup{
  colorlinks=true,
  linkcolor=blue,
  citecolor=ForestGreen,
  urlcolor=DarkOrchid
}

\AtBeginDocument{
  
}

\makeatother

\usepackage{babel}

\begin{document}

\title{Graph Coloring with Physics-Inspired Graph Neural Networks}

\author{Martin~J.~A.~Schuetz}
\affiliation{Amazon Quantum Solutions Lab, Seattle, Washington 98170, USA}
\affiliation{AWS Intelligent and Advanced Compute Technologies, 
Professional Services, Seattle, Washington 98170, USA}
\affiliation{AWS Center for Quantum Computing, Pasadena, CA 91125, USA}

\author{J.~Kyle~Brubaker}
\affiliation{AWS Intelligent and Advanced Compute Technologies, 
Professional Services, Seattle, Washington 98170, USA}

\author{Zhihuai~Zhu}
\affiliation{AWS Intelligent and Advanced Compute Technologies, 
Professional Services, Seattle, Washington 98170, USA}

\author{Helmut G.~Katzgraber}
\affiliation{Amazon Quantum Solutions Lab, Seattle, Washington 98170, USA}
\affiliation{AWS Intelligent and Advanced Compute Technologies, 
Professional Services, Seattle, Washington 98170, USA}
\affiliation{AWS Center for Quantum Computing, Pasadena, CA 91125, USA}

\date{\today}

\begin{abstract}

We show how graph neural networks can be used to solve the canonical
graph coloring problem. We frame graph coloring as a multi-class node
classification problem and utilize an unsupervised training strategy
based on the statistical physics Potts model.  Generalizations to other
multi-class problems such as community detection, data clustering, and
the minimum clique cover problem are straightforward. We provide
numerical benchmark results and illustrate our approach with an
end-to-end application for a real-world scheduling use case within a
comprehensive encode-process-decode framework.  Our optimization approach performs on
par or outperforms existing solvers, with the ability to scale to
problems with millions of variables.

\end{abstract}

\date{\today}

\maketitle

\section{Introduction}
\label{Introduction}

The graph coloring problem (GCP) is arguably one of the most famous
problems in the field of graph theory \citep{lewis:16, garey:79}.
Phrased as an optimization problem, the goal is to find an assignment of
labels (traditionally referred to as \textit{colors}) to the vertices
(nodes) of a graph such that no two adjacent vertices are of the same
color, while using the smallest number of colors possible.  The
convention of using colors dates back to the historic inception of this
problem: trying to color a map of the counties of England with the
smallest number of colors sufficient to color the map such that no
regions sharing a common border would be assigned the same color
\citep{lewis:16}. Today graph coloring is still an active field of
research, with real-world applications across a strikingly wide range of
domains, including (for example) the production of sports schedules, the
assignment of taxis to customer requests, the creation of timetables at
schools and universities, the allocation of computer programming
variables to computer registers, air traffic flow management
\citep{barnier:04}, and the game of Sudoku, among others
\citep{lewis:16}.

With online access to first-generation quantum computers steadily
expanding, the GCP has recently attracted considerable interest in the
broader quantum computing community.  In the current era of noisy
intermediate-scale quantum (NISQ) devices, typical approaches either
involve hybrid quantum-classical algorithms such as the Quantum
Approximate Optimization Algorithm (QAOA) \citep{oh:19} or quantum
annealing \citep{titiloye:11, pokharel:21}.  Given the low-level access
to these devices, the GCP typically has to be cast as a quadratic
unconstrained binary optimization problem (QUBO) \citep{glover:18} or,
equivalently, as an Ising Hamiltonian \citep{lucas:14}, at the expense
of increased resource requirements. Specifically, the QUBO description
of the GCP with $q>2$ colors for a graph with $n$ nodes requires $q
\times n$ binary variables, or (logical) qubits in the corresponding
quantum-native or quantum-inspired approach.  In addition, the
constraint that each vertex is assigned exactly one color has to be
enforced by hand with additional penalty terms \citep{lucas:14}.
Because of this added overhead due to the binary representation, it
would be preferable to tackle the problem in its native mathematical
form. In this work we propose the use of graph neural networks to do so,
aided by statistical physics concepts. 

\begin{figure}[b]
\includegraphics[width=1.0 \columnwidth]{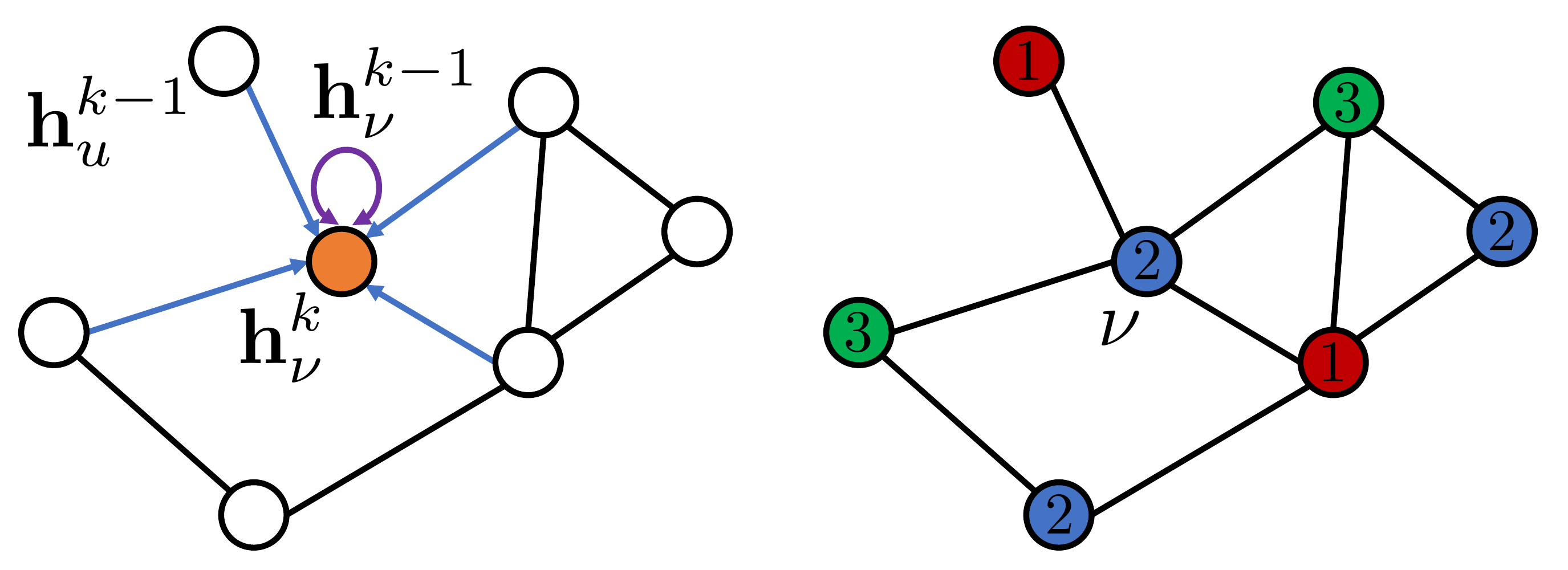}
\caption{
%(Color online) 
Schematic illustration of our approach. Following a recursive
neighborhood aggregation scheme, the graph neural network is iteratively
trained against a loss function based on the Potts model (enforcing
different color assignments to adjacent nodes). At training completion,
the final values for the soft node assignments at the final graph neural
network layer are projected to hard class (color) assignments
$\sigma_{i}=1,\dots, q$, as illustrated here for $q=3$ colors.  This
solution is optimal as the sample graph contains maximum cliques of size
three.
\label{fig:scheme}}
\end{figure}

In the deep learning community, graph neural networks (GNNs) have
emerged as a novel class of neural network architectures designed to
consume graph structure data \citep{gori:05, scarselli:08, micheli:09,
duvenaud:15, hamilton:17, xu:19, kipf:17, wu:19}, with the ability to
learn effective feature representations of nodes, edges, or even entire
graphs. Paradigmatic problems studied with GNNs can be categorized as
node classification, link prediction, graph classification, or community
detection, among others.  Prime examples include the classification of
users in social networks \citep{perozzi:14, sun:18}, the prediction of
future interactions in recommender systems \citep{ying:18}, and the
prediction of certain properties of molecular graphs \citep{strokach:20,
gaudelet:20}.  Leaving the details of specific GNN implementations aside
(see Refs.~\citep{li:16, velickovic:18, kipf:17} for further details),
the underlying theme for GNNs is the implementation of a \textit{message
passing} \citep{gilmer:17} scheme whereby GNNs iteratively update the
node (or edge) embeddings by aggregating information from their local
neighbors following the topology of the underlying graph.  Because of
their inherent scalability and graph-based design, GNNs present a
platform that can solve the graph coloring problem at scale.  We have
previously presented a physics-inspired, GNN-based framework to
(approximately) solve quadratic unconstrained binary \cite{glover:18}
combinatorial optimization problems with up to millions of variables
\citep{schuetz:21}.  In this work we natively extend this framework to
multi-color decision variables, and show how to solve the graph coloring
problem (GCP) without the need for extra penalty terms as needed when
using a QUBO-based approach. To this end we frame the GCP as a
multi-class node classification problem and use an unsupervised training
strategy based on the Potts model \cite{wu:82}, a generalization of the
Ising model in statistical physics.  For illustration purposes, our
approach is schematically depicted in Fig.~\ref{fig:scheme}. As
discussed in more detail below, generalizations of this approach to
applications such as data clustering \citep{blatt:96} and community
detection \citep{fortunato:10, fortunato:16, newman:04} are
straightforward.

The paper is structured as follows. In Sec.~\ref{Related Work} we
provide some context for our work, discussing the relevant literature at
the cross-section between graph coloring and graph neural networks.  In
Sec.~\ref{Preliminaries} we describe the basic concepts for our work,
with details on the GCP, the physics of the Potts model and its inherent
connection to the GCP, and graph neural networks.  In Sec.~\ref{GNN} we
then detail the theoretical framework underlying our approach, providing
a comprehensive physics-inspired, GNN-based approach towards solving the
GCP and related multi-color optimization problems.  Section
\ref{Application} outlines an end-to-end application for a real-world
scheduling use case, followed by numerical experiments in
Sec.~\ref{Numerics}.  Finally, in Sec.~\ref{Conclusion} we draw
conclusions and give an outlook on future directions of research.

\section{Related Work} 
\label{Related Work}

In this section we briefly review relevant existing literature, with the
goal to provide a detailed context for our work.  To keep the scope
manageable, we focus on work using GNN-based solution strategies.  For
an extensive review of the GCP we refer to Ref.~\citep{lewis:16}.

\textbf{Supervised Learning.} In Ref.~\citep{lemos:19} the authors
devise a binary classifier to solve the \textit{decision} version of the
graph coloring problem, {\em i.e.}, whether or not a given graph is
$q$-colorable.  To this end, Lemos \textit{et al.}~propose a model that
combines a graph neural network with a multi-layer perceptron. To train
this model a standard binary cross entropy loss function is used,
comparing the model's final prediction with the known ground-truth for a
given GCP instance, as obtained with a complementary CSP solver, albeit
for small problem instances only that can be solved exactly. As
discussed in Ref.~\citep{karalias:20}, such a supervised approach
critically depends on the existence of representative, labelled training
data sets with previously optimized hard problem instances, resulting in
a somewhat problematic chicken-and-egg scenario. In contrast to our
model, the approach outlined by Lemos \textit{et al.}~can underestimate
the chromatic number, and---going beyond binary graph
classification---requires a heuristic clustering algorithm such as
$k$-means in order to provide a constructive coloring solution.

\textbf{Unsupervised Learning.} Conceptually, our work is most similar
to those approaches that aim to train neural networks in an
unsupervised, end-to-end fashion, without the need for labelled training
sets.  Specifically, Li \textit{et al.}~have recently used graph neural
networks to solve the GCP following an unsupervised training strategy
\citep{li:21}. With a focus on the formal discriminative power of GNNs
for the graph coloring problem and motivated by mere intuition the
authors utilize a loss function which, as we show in this work, follows
straightforwardly from the Potts model, and therefore emerges as part of
a larger, unifying, \textit{physics-inspired} framework.  We also find
that our solver improves upon the results of Li \textit{et al.}~on
several benchmark problems.

Against this background, our work makes a \textit{physics-inspired}
contribution to the emerging cross-fertilization between combinatorial
optimization and machine learning \citep{kotary:21, cappart:21}.
Specifically, we provide a unified framework that pairs the Potts model
\cite{wu:82}---as extensively studied in the context of statistical
physics---with deep learning tools in the form of graph neural networks
to model and solve a large class of graph-based, multi-color
optimization problems such as graph coloring, community detection or
data clustering, all within a completely unsupervised, end-to-end
framework.

\section{Preliminaries}
\label{Preliminaries}

To set up our notation and terminology, we first provide a formal
problem definition for the graph coloring problem (GCP).  We then
highlight its close connection to the Potts model. Finally, we provide a
brief review of graph neural networks.

\textbf{Graph coloring.} We consider an undirected graph
$\mathcal{G}=(\mathcal{V}, \mathcal{E})$ with vertex set
$\mathcal{V}=\{1, 2, \ldots, n\}$ and edge set $\mathcal{E}=\{(i,j):i, j
\in \mathcal{V}\}$.  Given such a graph, in the graph coloring problem
we seek to assign an integer $c(\nu) \in \{1, 2, \dots, q\}$ to every
vertex $\nu \in \mathcal{V}$, such that (i) the assignment is free of
color clashes, i.e., $c(u) \neq c(v) \hspace{0.1cm} \forall (u,v) \in
\mathcal{E}$, and (ii) the number of colors $q$ is minimal.  We refer to
a clash-free coloring using at most $q$ colors as a proper (feasible)
$q$-coloring. If such a $q$-coloring can be found, the graph is said to
be $q$-colorable. The chromatic number of a graph $\mathcal{G}$, denoted
as $\chi=\chi(\mathcal{G})$, with $1\leq \chi \leq n$, is the minimum of
colors required for a feasible coloring of $\mathcal{G}$.  Accordingly,
our goal is to find the chromatic number $\chi$, with a coloring where
adjacent vertices are assigned to different colors. In general, this
problem is computationally hard, with exact algorithms displaying an
exponential runtime in the size of the input $n$. Specifically, it is
known to be NP-hard to compute the chromatic number $\chi$, typically
leaving heuristics such as greedy coloring or tabu-search based methods
as the go-to approximation strategies \citep{lewis:16}.

The graph coloring problem is also closely related to yet another
NP-hard combinatorial optimization problem, the minimum clique cover
problem (MCC) \citep{karp:72}.  In MCC the goal is to partition the
nodes of a graph into cliques, with as few cliques as possible.
Conversely, graph coloring provides color classes, {\em i.e.},
partitions of the vertex set into independent sets (that is subsets with
no adjacencies), yielding the following equivalence between clique
covers and coloring: Because a subset of vertices is a clique in
$\mathcal{G}$ if and only if it is an independent set in the complement
of $\mathcal{G}$, a partition of the vertices of $\mathcal{G}$ is a
clique cover of $\mathcal{G}$ if and only if it is a coloring of the
complement of $\mathcal{G}$. For a given graph $\mathcal{G}$, the
smallest number for which a clique cover exists is called the clique
cover number.

\textbf{Potts model.} The GCP outlined above is closely related to the
standard Potts model, as argued below.  In the Potts model every vertex
is associated with a spin variable $\sigma_{i}=1, \dots, q$ that can
take on $q$ different values. The Hamiltonian for the
Potts model can be expressed in compact form as
\begin{equation}
H_\mathrm{Potts}  =  -J \sum_{(i,j)\in \mathcal{E}} \delta(\sigma_{i}, \sigma_{j}), \label{eq:Potts-Hamiltonian}
\end{equation}
where $\delta(\sigma_{i}, \sigma_{j})$ refers to the Kronecker delta,
which equals one whenever $\sigma_{i} = \sigma_{j}$ and zero otherwise,
thus capturing the hard-core spin-spin interactions characteristic for
the Potts model \citep{wu:82, zdeborova:07}.  Accordingly, if two
adjacent spins $\sigma_{i}$ and $\sigma_{j}$ are in the same state, the
energy contribution is $-J$, while it is zero whenever they are in
different states. To enforce a feasible coloring of the underlying
graph, we consider anti-ferromagnetic interactions and (if not stated
otherwise) set $J=-1$ in the following. 

Generalizations to settings with
weighted interactions $J_{ij}$ (where some constraints are more
important than others) are straightforward.  Important applications
thereof include data clustering \citep{blatt:96} and community detection
as captured by the maximization of the modularity parameter
\citep{fortunato:10, fortunato:16, newman:04}, among others.
Specifically, we find that the latter can be described by the
generalized Potts Hamiltonian
\begin{equation}
H_\mathrm{Potts}  =  -\sum_{i,j} J_{ij}\delta(\sigma_{i}, \sigma_{j}), \label{eq:Potts-Hamiltonian-community}
\end{equation}
with interaction strength
\begin{equation}
J_{ij}=\frac{1}{2m}\left(A_{ij}-\frac{d_{i}d_{j}}{2m}\right), 
\end{equation}
where $A_{ij}$ refers to the adjacency matrix of the graph,
$d_{i}=\sum_{j}A_{ij}$ is the degree of node $i$, and $m=1/2
\sum_{i}d_{i}$.  Finally, for two colors ($q=2$) the Potts model reduces
to the MaxCut problem, with Hamiltonian
$H_{\mathrm{MaxCut}}=\sum_{i<j}J_{ij}z_{i}z_{j}$ with $J_{ij}=A_{ij}/2$
and binary spin variables $z_{i}\in\{-1,1\}$ \citep{schuetz:21}, as can
be seen by the transformation $\delta(\sigma_{i}, \sigma_{j})
\rightarrow (1+z_{i}z_{j})/2$.

The close connection between the GCP and the Potts model becomes
apparent in the (dimensionless) partition function \cite{yeomans:92} of
the Potts model, which allows one to compute most
thermodynamic variables of a system through derivatives.  
For the Potts model it is given by 
\begin{equation}\mathcal{Z} =
\sum_{\{\sigma_{i}\}}\exp[-\beta H_\mathrm{Potts}],
\end{equation}
where $\beta=1/k_{B}T$ is the inverse temperature (with $T>0$) and
$k_{B}$ is the Boltzmann constant.  Using the relation $\exp[K
\delta(\sigma_{i}, \sigma_{j})] = 1+[\exp(K)-1] \delta(\sigma_{i},
\sigma_{j})$, we then find in generality
\begin{equation}
\mathcal{Z} = \sum_{\{\sigma_{i}\}} \prod_{(i,j)\in
\mathcal{E}} \left[1+[\exp(K)-1] \delta(\sigma_{i}, \sigma_{j})
\right] .  
\end{equation}
In the zero-temperature limit we obtain \cite{wu:88} 
\begin{equation}
\mathcal{Z}
\rightarrow P_{\mathcal{G}}(q) = \sum_{\{\sigma_{i}\}} \prod_{(i,j)\in
\mathcal{E}} \left[1-\delta(\sigma_{i}, \sigma_{j}) \right]
\;\;{\rm for}\;\;
T \to 0.
\end{equation}
In the last step we have introduced the chromatic function (polynomial)
$P_{\mathcal{G}}(q)$, a central quantity in the theory of graph
coloring, thus directly relating the Potts model to graph coloring.  In
the limit $T \rightarrow 0$ adjacent spins are forced to occupy
different states, and the partition function $\mathcal{Z}$ simply
reduces to the chromatic function $P_{\mathcal{G}}(q)$ which counts the
number of possible $q$-colorings of $\mathcal{G}$ as a function of the
number of colors $q$.  The chromatic number $\chi = \min \left\{ q \in
\mathbb{N}: P_{\mathcal{G}}>0 \right\}$ is then the smallest positive
integer that is not a zero of the chromatic polynomial.

\textbf{Graph Neural Networks.} Graph neural networks are an emergent
family of neural networks that extend the standard deep learning toolbox
to graph data \citep{hamilton:20}.  While convolutional neural networks
are well-defined only over rigid, grid-structured data (such as images),
and recurrent neural networks are built for sequences of data (such as
text), the GNN formalism provides a general framework for defining
neural networks on graph-structured data \citep{hamilton:20}.  With
permutation invariance (under the arbitrary labeling of nodes) built in
by design, GNNs offer a scheme to generate node representations that
incorporate the topology of the graph.  The common theme to any type of
GNN is that it implements some form of neural \textit{message passing},
whereby messages (in the form of vectors) are exchanged between the
nodes of the graph to iteratively update the internal representations of
the graph's nodes \citep{gilmer:17}.  More formally, for a given input
graph $\mathcal{G}=(\mathcal{V}, \mathcal{E})$ along with any relevant
node features $\mathbf{X} \in \mathbb{R}^{d_{0} \times n}$, a GNN can be
used to generate node embeddings $\mathbf{p}_{\nu}, \forall \nu \in
\mathcal{V}$ \citep{hamilton:20}.  This is done iteratively as follows:
Consider hidden embedding vectors $\{\mathbf{h}_{\nu}^{k}\}$
representing each node $\nu \in \mathcal{V}$.  In each iteration $k$,
every embedding vector $\mathbf{h}_{\nu}^{k}$ is updated based on
information inferred from the corresponding local neighborhood, denoted
as $\mathcal{N}_{\nu} = \{ u \in \mathcal{V} | (u,\nu) \in \mathcal{E}
\}$. At layer (iteration) $k=0$, the initial representations
$\mathbf{h}_{\nu}^{0} \in \mathbb{R}^{d_{0}}$ are usually derived from
the node's labels or given input features of dimensionality $d_0$
\citep{alon:21}.  This single-layer update can then be formalized as
\begin{eqnarray}
\mathbf{m}_{\nu}^{k} &=& \mathsf{AGGREGATE}_{\theta}^{k} \left( \{\mathbf{h}_{u}^{k-1} | u \in \mathcal{N}_{\nu} \} \right), \nonumber  \\
\mathbf{h}_{\nu}^{k} &=& \mathsf{UPDATE}_{\theta}^{k} \left(\mathbf{h}_{\nu}^{k-1},  \mathbf{m}_{\nu}^{k} \right),  \label{eq:GNN}
\end{eqnarray}
for the GNN layers (iterations) $k=1, \dots, K$, with
$\mathsf{AGGREGATE}(\cdot)$ and $\mathsf{UPDATE}(\cdot)$ referring to
some (typically parametrized) differentiable functions
\citep{hamilton:20}. See Refs.~\citep{li:16, velickovic:18, kipf:17,
wu:19} for several popular design choices such as graph convolutional
networks (GCNs). In Eq.~\eqref{eq:GNN}, the vector
$\mathbf{m}_{\nu}^{k}$ represents the $k$-th layer \textit{message}  for
node $\nu =1, \dots, n$ as aggregated from the corresponding local graph
neighborhood $\mathcal{N}_{\nu}$.  At each iteration $k$, every node
aggregates information from its local neighborhood, and as these
iterations progress each node embedding encapsulates a larger receptive
field within the graph.  Specifically, after $k$ iterations every node
embedding contains information about its $k$-hop neighborhood, with the
final output (after $K$ iterations of message passing) defined as
$\mathbf{p}_{\nu} = \mathbf{h}_{\nu}^{K}$.  This output can then be used
for prediction tasks, such as node classification.  To optimize the
predictive power of this approach, the (parametrized) final node
embeddings  $\mathbf{p}_{\nu} = \mathbf{h}_{\nu}^{K}(\theta)$ are fed
into a problem-specific loss function, with some form of stochastic
gradient descent optimizing the weight parameters $\theta$ of the
network.

\section{Theoretical framework}
\label{GNN}

\begin{figure}
\includegraphics[width=1.0 \columnwidth]{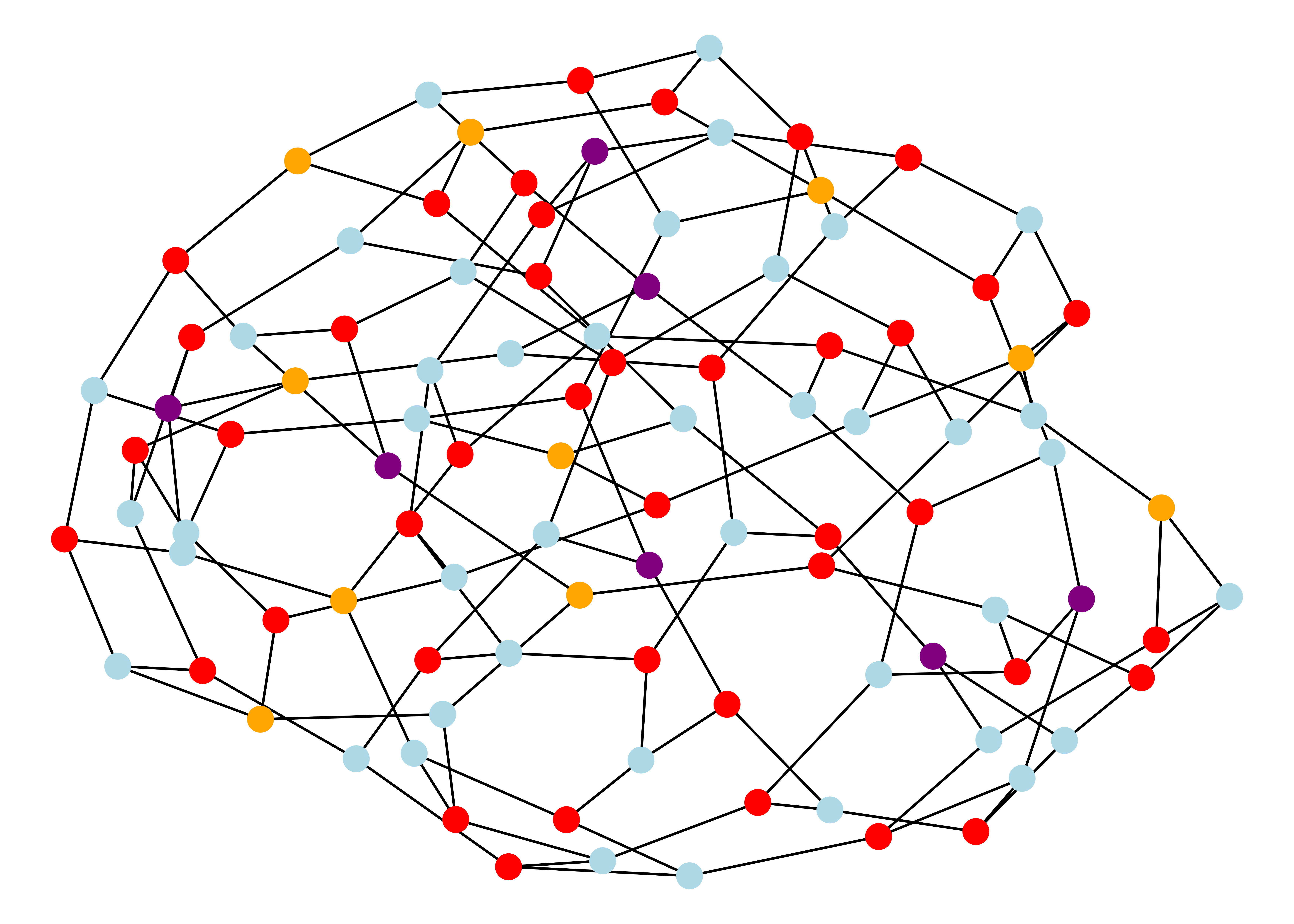}
\caption{
%(Color online) 
Example $4$-coloring solution to the graph coloring problem for a random
$3$-regular graph with $n=100$ nodes.  At training completion the GNN
provides color (class) assignments to each vertex.  The optimization
problem is to assign the colors in a way that adjacent nodes must be
assigned different colors, while using the smallest number of colors
possible (corresponding to the antiferromagnetic ground-state of the
underlying Potts model).
\label{fig:graph-example-solution}}
\end{figure}

\begin{figure*}
\includegraphics[width=1.8 \columnwidth]{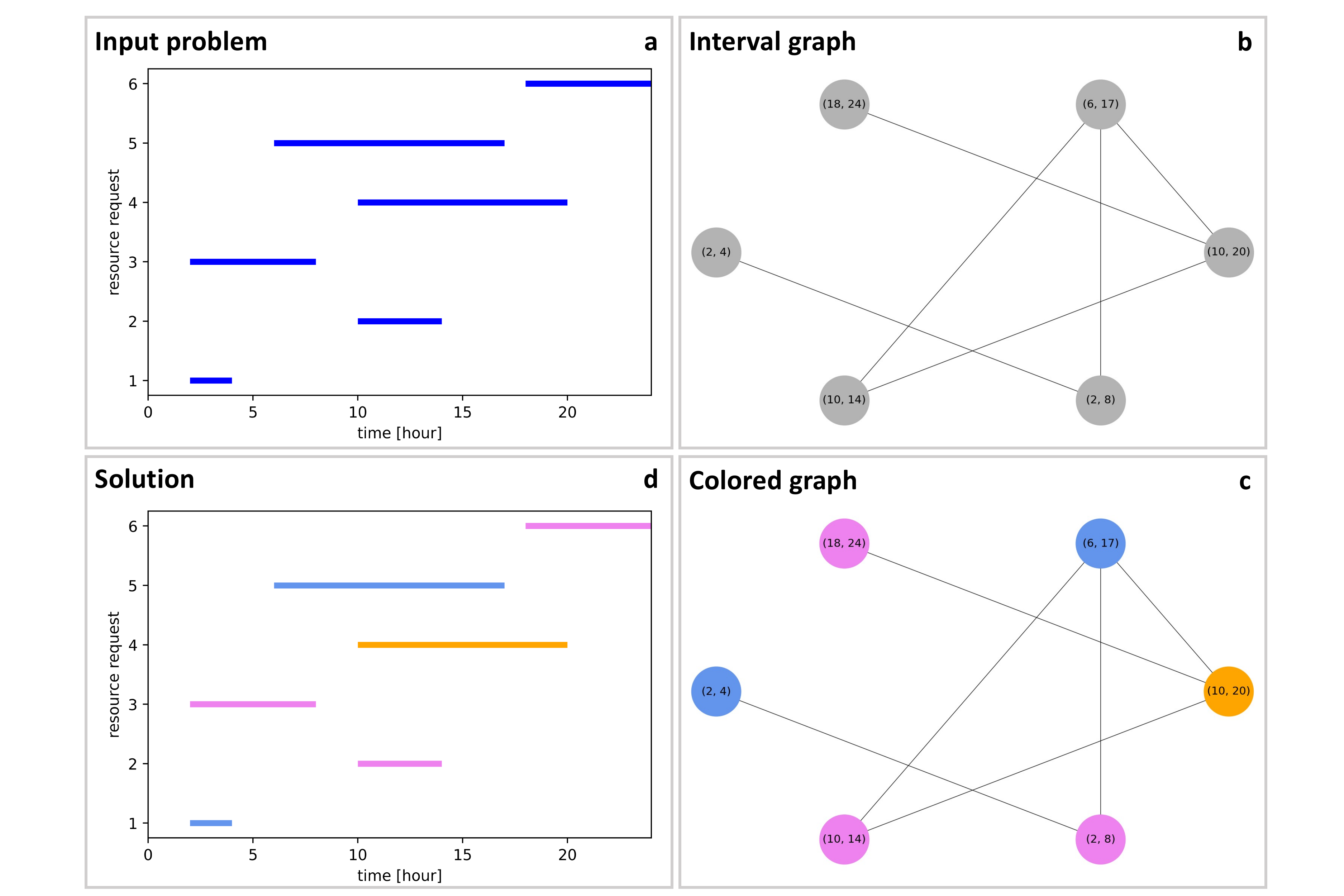}
\caption{
%(Color online) 
Example end-to-end application of graph coloring for a task scheduling
problem. \textbf{(a)}, The problem is specified in terms of a schedule
detailing six resource requests (vertical axis) as a function of time,
spread out over the course of 24 hours (horizontal axis).  \textbf{(b)},
\textit{Encoding}: The problem is encoded in the form of an
\textit{interval graph} where every node represents one request labelled
by the corresponding time interval, and edges refer to clashes within
the resource requests whenever two requests overlap in time.
\textbf{(c)}, \textit{Processing}: We solve the graph coloring problem
on this interval graph using a graph neural network with a Potts-type
loss function as detailed in the main text.  Once the algorithm has
converged, we obtain a graph colored with the smallest number of color
clashes for the given number of colors.  In this example we find a
feasible coloring with $\chi=3$ colors as expected based on the clique
of size three.  \textbf{(d)}, \textit{Decoding}: Finally the proposed
colors are mapped back to the original resource requests. In this
example we find that three resources are sufficient in order to satisfy
all requests.\label{fig:application}
}
\end{figure*}

In this section we discuss in detail the theoretical framework
underlying our work.  We show how to solve the GCP using GNNs, with the
anti-ferromagnetic Potts model providing a canonical choice for the loss
function controlling the unsupervised GNN training process.

We consider an undirected graph $\mathcal{G}=(\mathcal{V}, \mathcal{E})$
with vertex set $\mathcal{V}=\{1, 2, \ldots, n\}$ and edge set
$\mathcal{E}=\{(i,j):i, j \in \mathcal{V}\}$.  Given such graph, our
goal is to assign colors to the nodes of the graph in such a way that
adjacent nodes are assigned different colors and the number of colors
used is minimal.  To this end we associate a discrete variable (spin)
$\sigma_{\nu}=1, \dots, q$ with every vertex $\nu \in \mathcal{V}$,
thereby assigning one of $q$ possible states (colors) to every node in
the graph.  To enforce a valid coloring we consider the standard Potts
spin model \citep{wu:82} with anti-ferromagnetic interactions as given
in Eq.~\eqref{eq:Potts-Hamiltonian}; this model gives no energy
contribution to neighboring spins with different colors, but penalizes
color clashes with a positive energy offset.  The ground-state energy is
then zero if and only if the graph is $q$-colorable, thus providing a
good cost function for encoding the GCP. To make the GCP compatible with
our GNN-based approach we first reformulate the Potts model
\eqref{eq:Potts-Hamiltonian} in terms of one-hot-encoded variables
$\hat{\mathbf{y}}_{i}$ as
\begin{equation}
H_\mathrm{Potts}  =  -J \sum_{(i,j)\in \mathcal{E}} \hat{\mathbf{y}}_{i}^{\intercal} \cdot \hat{\mathbf{y}}_{j}.
\end{equation}  
Here, the variable $\hat{\mathbf{y}}_{i}$ describes the class assignment
for node $i \in \mathcal{V}$ within a $q$-dimensional unit vector where
all components are zero except for one (set to 1) and signals the color
assignment as $\sigma_{i} = \sum_{\alpha=1}^{q}\alpha
\hat{\mathbf{y}}_{i}^{[\alpha]}$, with $\hat{\mathbf{y}}_{i}^{[\alpha]}$
denoting the $\alpha$-th component of $\hat{\mathbf{y}}_{i}$, and by
definition $\sum_{\alpha}\hat{\mathbf{y}}_{i}^{[\alpha]}=1$.  Next,
generalizing our approach as detailed in Ref.~\citep{schuetz:21} to
multi-class node classification problems, we apply a relaxation strategy
to the problem Hamiltonian $H_\mathrm{Potts}$ to generate a
differentiable loss function $\mathcal{L}(\theta)$ with which we perform
unsupervised training on the multi-color node representations of the
GNN.  To this end we replace the (hard) one-hot-encoded decision vectors
$\hat{\mathbf{y}}_{i}$ with corresponding (soft) normalized assignments
$\mathbf{p}_{i}(\theta) \in [0,1]^{q}$, letting $\hat{\mathbf{y}}_{i}
\longrightarrow \mathbf{p}_{i}(\theta)$.  In our approach, these soft
assignments $\mathbf{p}_{i}(\theta)$ are generated by our GNN Ansatz as
final node embeddings $\mathbf{p}_{i} = \mathbf{h}_{i}^{K} \in
[0,1]^{q}$ at layer $K$, after the application of a standard softmax
activation function, and used as an input for the Potts-like loss
function $\mathcal{L}(\theta)$ given by
\begin{equation}
H_\mathrm{Potts} \longrightarrow \mathcal{L}_\mathrm{Potts} (\theta) = -J \sum_{(i,j) \in \mathcal{E}} \mathbf{p}_{i}^{\intercal} \cdot \mathbf{p}_{j}.  \label{eq:loss-Potts}
\end{equation} 
To arrive at the predicted soft assignments $\mathbf{p}_{i}$ for all
nodes $i=1, \dots, n$, the GNN follows a standard recursive neighborhood
aggregation scheme \citep{gilmer:17, xu:18}, where each node $\nu=1, 2,
\ldots, n$ collects information (encoded as feature vectors) of its
neighbors to compute its new feature vector $\mathbf{h}_{\nu}^{k}$ at
layer $k=0, 1, \ldots, K$. Similar to Ref.~\citep{schuetz:21}, the node
embeddings $\mathbf{h}_{\nu}^{0}$ are initialized randomly.  After $k$
iterations of aggregation, a node is represented by its transformed
feature vector $\mathbf{h}_{\nu}^{k}$, which captures the structural
information within the node's $k$-hop neighborhood \citep{xu:19}.  For
the multi-class node classification task at hand we use convolutional
aggregation steps, followed by the application of a nonlinear softmax
activation function with the dimensionality set by the number of colors
$q$, thereby providing one-hot-encoded $q$-dimensional soft
(probabilistic) node assignments $\mathbf{p}_{\nu} =
\mathbf{h}_{\nu}^{K} \in [0,1]^{q}$, with the softmax function
automatically ensuring normalization as
$\sum_{\alpha=1}^{q}\mathbf{p}_{\nu}^{[\alpha]}=1$.  By virtue of this
built-in normalization and in stark contrast to any QUBO-based approach
\citep{glover:18, lucas:14, kochenberger:05}, we do not have to add
additional terms to the loss function to enforce a one-hot constraint
that drives the solution towards one \textit{unique} color assignment
per node. Conversely, once the unsupervised training process has
completed, we apply a simple projection heuristic to map the soft
assignments $\mathbf{p}_{\nu}$ to hard class variables $\sigma_{\nu}=1,
\dots, q$ using, for example,
$\sigma_{\nu}=\mathrm{argmax}(\mathbf{p}_{\nu})$ to find the class
(color) with the largest predicted probability, thus providing unique
color assignments for every node.  As shown in Fig.~\ref{fig:scheme},
the final color assignment $\{\sigma_{\nu}\}$ can then be visualized as
a $q$-coloring of the graph.  For further illustration, an example
$4$-coloring solution (as implemented with this approach) for a random
$3$-regular graph with $n = 100$ vertices is shown in
Fig.~\ref{fig:graph-example-solution}.  Far beyond this sample scale,
the scalability inherent to GNNs opens up the possibility of studying
unprecedented problem sizes with hundreds of millions of nodes when
leveraging distributed training in a mini-batch fashion on a cluster of
machines as demonstrated recently in Ref.~\citep{zheng:20}.

Our approach features several hyperparameters, including the number of
layers $K$, the dimensionality of the embedding vectors
$\mathbf{h}_{i}^{k}$, and the learning rate $\beta$, which can be
optimized via hyperparameter optimization techniques.  In particular,
the number of colors $q$ can be seen as a GCP-specific hyperparameter
that together with the graph $\mathcal{G}$ defines the input pair
$(\mathcal{G}, q)$ for the decision problem whether or not $\mathcal{G}$
allows for a $q$-coloring.  To identify the chromatic number $\chi$, one
can perform, for example, a naive search by sequentially checking if
$\mathcal{G}$ is $q$-colorable for $q=1, 2, \dots$, or use a binary
search to cut down the average number of calls required logarithmically.
Alternatively, one can try to solve the graph coloring problem ({\em
i.e.}, the search for a feasible coloring) in parallel with the
minimization of colors used by adding a corresponding regularization
term to the loss function (such as $\sim q^2$).  However, such term
should not overpower the regular Potts-like term
$\mathcal{L}_\mathrm{Potts} (\theta)$, as to not drive the overall
solution towards an infeasible coloring.

\section{Industry Applications}
\label{Application}

The graph coloring problem is known to describe many real-world
applications, in particular in scheduling and allocation problems
\citep{lewis:16}.  Prominent examples include timetabling problems or
frequency assignment problems, relevant to the planning of wireless
communication services \citep{lewis:16, eisenblaetter:02}.  To
illustrate both our GNN-based approach as well as the real-world
applicability of the graph-coloring problem, we now discuss an
end-to-end application for a canonical scheduling use case.  We do so
within a comprehensive three-step encode-process-decode approach in
which we (i) first phrase the use case as a graph coloring problem
(encoding), (ii) we then solve this problem using our GNN-based approach
(processing), and finally (iii) decode the coloring solution to an
actual solution for the use case at hand (decoding).  For the sake of
this illustrative example we consider a small problem instance as
illustrated in Fig.~\ref{fig:application}. More thorough
numerical benchmarks are presented in Sec.~\ref{Numerics}.

We consider a scenario involving the scheduling of tasks with given
start and end times, with applications in car-sharing, taxi companies,
aircraft assignments, etc.  Specifically, we face $n$ resource requests
(or bookings) with a start time indicating when the resource will be
needed and an end time indicating the resource is available; see
Fig.~\ref{fig:application}(a) for an example problem with $n=6$ resource
requests. The problem is then to assign resources ({\em e.g.}, cars) to
these requests ({\em e.g.}, bookings) in the most efficient way
involving the smallest number of resources needed. As illustrated in
Fig.~\ref{fig:application}(a), typically some requests will overlap in
time leading to request clashes that cannot be satisfied by the same
resource.  As commonly done in resource allocation problems and
scheduling theory, this situation can conveniently be described with the
help of an undirected \textit{interval graph} in which a vertex is introduced 
for every request, with edges connecting vertices whose requests overlap. 
Figure \ref{fig:application}(b) displays an encoding of the problem with a graph 
made of six vertices and six edges, including a clique of size
three.  While inexpensive, special-purpose algorithms exist for interval
graphs \citep{lewis:16}, we can then solve the graph coloring problem on
this interval graph (in the same way as any other GCP) using our
general-purpose GNN-based approach. To this end, we run unsupervised
multi-class classification directly on the interval graph with $n=6$
nodes and final softmax non-linearity of dimension $q=3$, as opposed to
QUBO-based approaches involving $q \times n=18$ binary variables
\citep{oh:19}. For the sample problem illustrated in
Fig.~\ref{fig:application}(c) we obtain a feasible coloring using just
three colors ($\chi=3$).  Finally, as shown in
Fig.~\ref{fig:application}(d), we decode this coloring to the
corresponding assignment in which three resources are used to satisfy
all six requests.

\section{Numerical Experiments} \label{Numerics}

\begin{table*}
\centering
 \begin{tabular}{| l | c c c | c | c | c | c c c c c | c |} 
 \hline
 \hline
 graph & nodes & edges & density & colors $q$ & $\bar{\chi}_{\mathrm{greedy}}$ & $\bar{\chi}_{\mathrm{GNN}}$ & Tabucol & Tabucol \citep{li:21} & GNN \citep{li:21} & PI-GCN & PI-SAGE & error $\epsilon$ \\ [0.5ex] 
 \hline\hline
 anna & 138 & 493 & 5.22\% & 11 & 11 & 11 & \textbf{0} & \textbf{0} & 1 & 1 & \textbf{0} & 0.00\%\\ 
 jean & 80 & 254 & 8.04\% & 10 & 10 & 10 & \textbf{0} & \textbf{0} & \textbf{0} & \textbf{0} & \textbf{0} & 0.00\%\\
 myciel5 & 47 & 236 & 21.83\% & 6 & 6 & 6 & \textbf{0} & \textbf{0} & \textbf{0} & \textbf{0} & \textbf{0} & 0.00\% \\ 
 myciel6 & 95 & 755 & 16.91\% & 7 & 7 & 7 & \textbf{0} & \textbf{0} & \textbf{0} & \textbf{0} & \textbf{0} & 0.00\% \\
 queen5-5 & 25 & 160 & 53.33\% & 5 & 5 & 5 & \textbf{0} & \textbf{0} & \textbf{0} & \textbf{0} & \textbf{0} & 0.00\% \\
 queen6-6 & 36 & 290 & 46.03\% & 7 & 8 & 7 & \textbf{0} & \textbf{0} & 4 & 1 & \textbf{0} & 0.00\% \\
 queen7-7 & 49 & 476 & 40.48\% & 7 & 9 & 7 & \textbf{0} & 10 & 15 & 8 & \textbf{0} & 0.00\% \\
 queen8-8 & 64 & 728 & 36.11\% & 9 & 10 & 10 & \textbf{0} & 8 & 7 & 6 & 1 & 0.14\% \\
 queen9-9 & 81 & 1056 & 32.59\% & 10 & 12 & 11 & \textbf{0} & 5 & 13 & 13 & 1 & 0.09\% \\
 queen8-12 & 96 & 1368 & 30.00\% & 12 & 13 & 12 & \textbf{0} & 10 & 7 & 10 & \textbf{0} & 0.00\% \\
 queen11-11 & 121 & 1980 & 27.27\% & 11 & 15 & 14 & 20 & 33 & 33 & 37 & \textbf{17} & 0.86\% \\
 queen13-13 & 169 & 3328 & 23.44\% & 13 & 17 & 17  & 35 & 42 & 40 & 61 & \textbf{26} & 0.78\%  \\ [1ex] 
 \hline
 \hline
 \end{tabular}
\caption{
Numerical results for COLOR graphs \citep{trick:02}.  For a given number
of colors $q$, we report the $\mathrm{cost} = H_{\mathrm{Potts}}$, that
is the number of conflicts in the best coloring result, as achieved with
our physics-inspired GNN solvers (PI-GCN and PI-SAGE), 
together with results for the Tabucol algorithm, as partially sourced from Ref.~\cite{li:21}. 
Upper bounds on the chromatic number $\chi$ as found by a greedy algorithm as well as PI-SAGE are reported as $\bar{\chi}_{\mathrm{greedy}}$ and $\bar{\chi}_{\mathrm{GNN}}$, respectively. 
Best results are marked in boldface. The last column
gives the normalized error $\epsilon$ (for the best PI-GNN result) specifying the relative fraction
of edges with color clashes. 
%Solutions marked with an asterisk $(^{*})$ are displayed in Fig.~\ref{fig:COLOR-graph}. 
Example solutions are displayed in Fig.~\ref{fig:COLOR-graph}.
Further details are provided in the main text.
\label{tab:color-graphs}} 
\end{table*}

\begin{figure}
\includegraphics[width=0.49 \columnwidth]{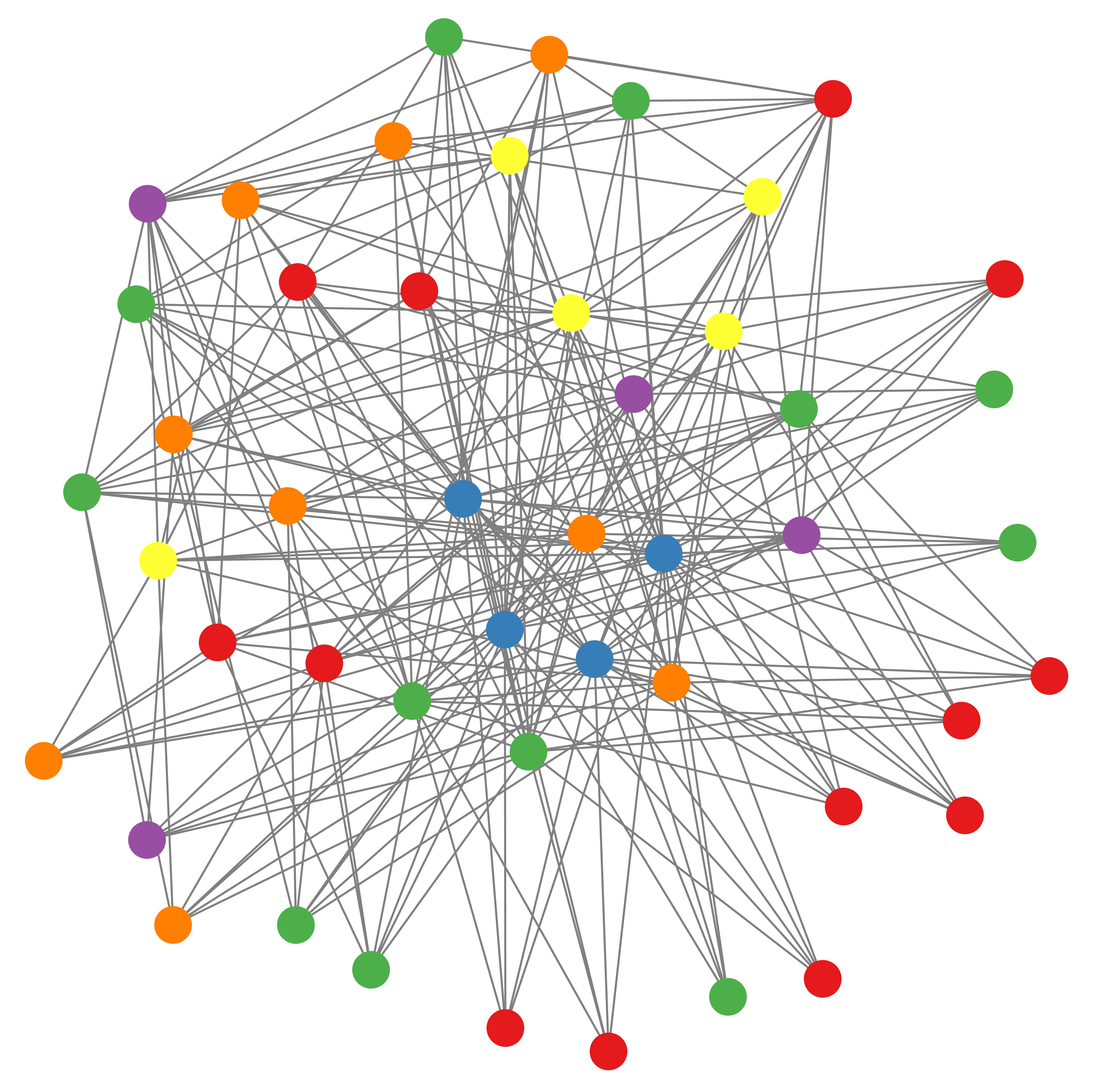}
\includegraphics[width=0.49 \columnwidth]{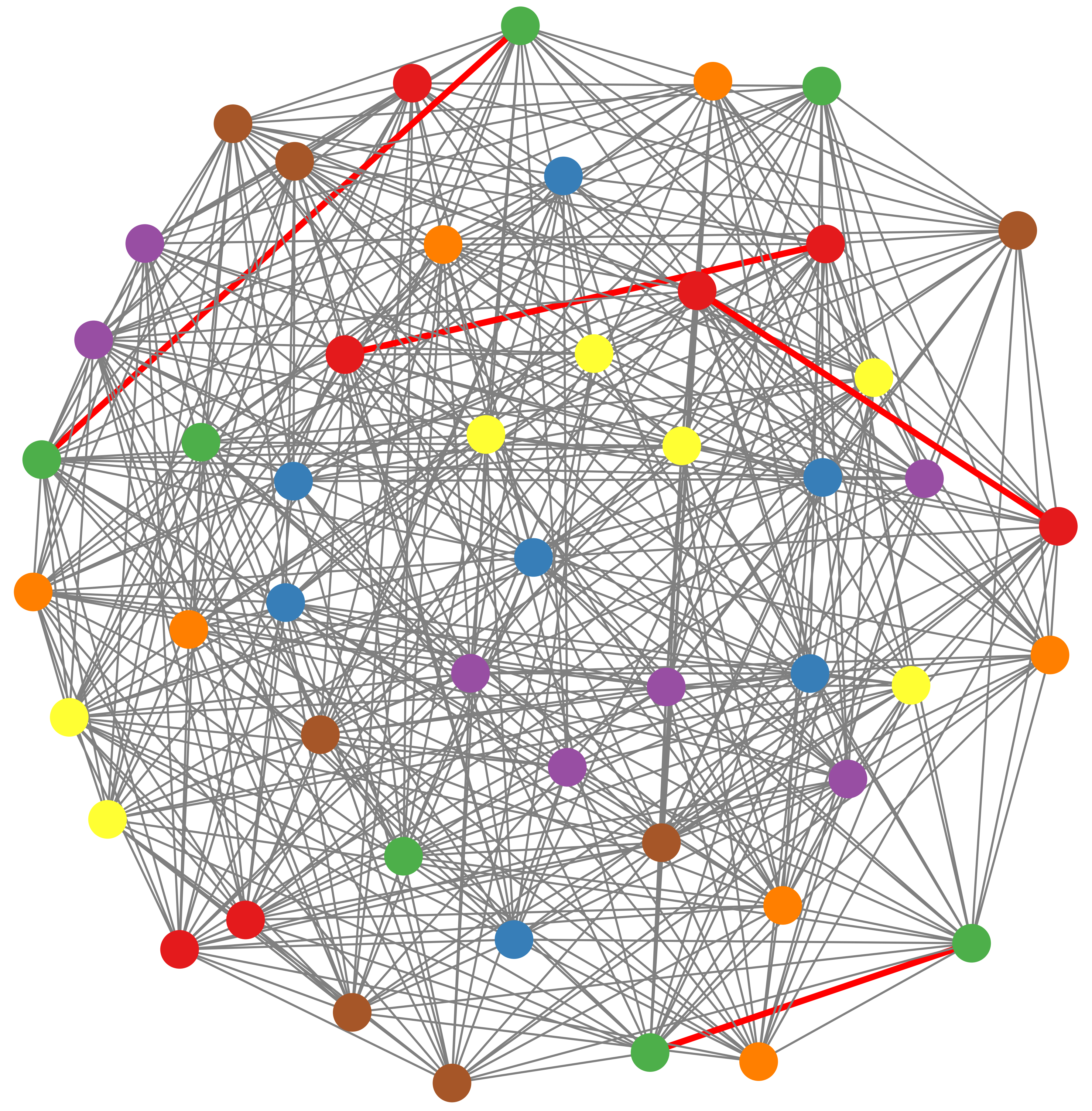}
\caption{
%(Color online) 
Example solutions to the graph coloring problem for the myciel5 (left) and
queen7-7 (right) graphs from the COLOR dataset with $q=6$ and $q=7$
colors, respectively. The solution for myciel5 corresponds to a feasible
coloring with normalized error $\epsilon=0\%$, whereas the solution for
queen7-7 represents an infeasible, but low-energy solution with
$\epsilon=0.84\%$ for which the remaining four color clashes have been
highlighted with bold (conflicting) edges.  Further details are provided
in Tab.~\ref{tab:color-graphs} and the main text.
\label{fig:COLOR-graph}
}
\end{figure}

\begin{table*}
\centering
 \begin{tabular}{| l | c c c | c | c | c | c c c c c | c |} 
 \hline
 \hline
 graph & nodes & edges & density & colors $q$ & $\bar{\chi}_{\mathrm{greedy}}$ & $\bar{\chi}_{\mathrm{GNN}}$ & Tabucol & Tabucol \citep{li:21} & GNN \citep{li:21} & PI-GCN & PI-SAGE & error $\epsilon$ \\ [0.5ex] 
 \hline\hline
 Cora & 2708 & 5429 & 0.15\% & 5 & 5 & 5 & \textbf{0} & 31 & 3 & 1 & \textbf{0} & 0.00\%\\ 
 Citeseer & 3327 & 4732 & 0.09\% & 6 & 6 & 6 & \textbf{0} & 6 & 3 & 1 & \textbf{0} & 0.00\%\\
 Pubmed & 19717 & 44338 & 0.02\% & 8 & 8 & 9 & NA & NA & 35 & \textbf{13} & 17 & 0.03\% \\ [1ex] 
 \hline
 \hline
 \end{tabular}
\caption{
Numerical results for citation graphs \citep{cora:00, citeseer:08,
pubmed:12}. Further details are provided in the caption of
Tab.~\ref{tab:color-graphs} and in the main text.
\label{tab:citation-graphs}} 
\end{table*}

\begin{figure}
\includegraphics[width=1.0 \columnwidth]{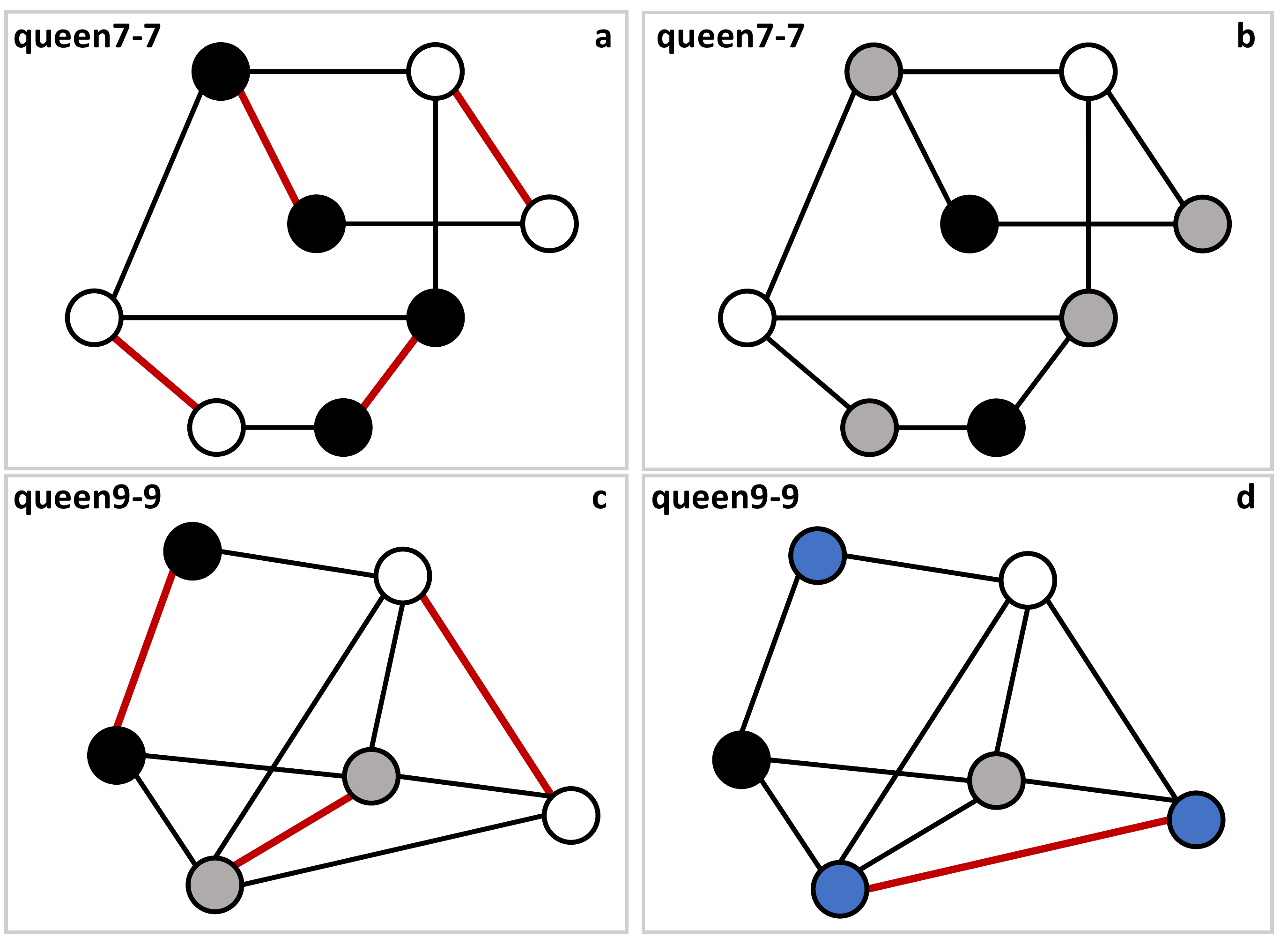}
\caption{
%(Color online) 
Illustration of our simple post-processing heuristic to estimate upper bounds on the chromatic number. 
Color clashes have been highlighted with bold (conflicting) edges. 
Panels on the left display example original (infeasible) solutions found by PI-GNN (not corresponding to the best solutions found with PI-GNN).
Panels on the right show the results of a simple post-processing strategy.
For the queen7-7 example this randomized post-processing heuristic can remove all color clashes and provide a feasible coloring, at the expense of one additional color. 
For the queen9-9 instance, one color clash is still outstanding after application of one iteration of this purification protocol. 
Further details are provided in the main text.
\label{fig:postprocessing}
}
\end{figure}

We now turn to systematic numerical experiments using standard benchmark
problems for graph coloring. In particular, we provide results for the
publicly available COLOR dataset \citep{trick:02}, as well as well-known
citation datasets (Cora \citep{cora:00}, Citeseer \citep{citeseer:08},
and Pubmed \citep{pubmed:12}) often used for graph-based benchmark
experiments. The former provide small and medium-sized dense problem
instances with relatively large known chromatic numbers ($\chi \sim
10$), while the latter are large, but sparse real-world graphs (which,
for the purpose of graph coloring, we consider as undirected graphs,
dismissing any potential node or edge features).  Our basic GNN
architecture is very similar to the one detailed in
Ref.~\citep{schuetz:21}, except for the dimension of the final GNN layer
set here to the number of colors $q$.  We provide results for two
standard types of GNN architectures, that is graph convolutional
networks (GCN) \citep{kipf:17} as well as GraphSAGE \citep{hamilton:17}.
Model configurations (hyperparameters) are detailed in the Supplemental
Material. We compare our results (as given by the cost directly reported
by the Potts Hamiltonian) to previously published results sourced from
Ref.~\citep{li:21}, including results based on the tabu-search based
heuristic called Tabucol \citep{hertz:87}, a local search algorithm
which tracks single moves within a tabu list.  
We complement these with our own benchmark results obtained with Tabucol and a \textit{greedy} coloring algorithm. 
The latter parses through the graph's vertices one by one according to some vertex ordering and greedily assigns the first available color. 
If no available color can provide a feasible coloring, yet another color is expensed, thus (by design) always providing a feasible coloring with a corresponding upper bound on the chromatic number denoted as $\bar{\chi}_{\mathrm{greedy}}$. 
Here, we have implemented a greedy algorithm with largest-first ordering strategy as further detailed in Refs.~\citep{lewis:16, kosowski:04}. 
Our greedy results for $\bar{\chi}_{\mathrm{greedy}}$ largely agree with results presented in Ref.~\citep{aslan:18}.

For a given graph and a
fixed number of colors $q$, we report the total number of color clashes
as achieved with our physics-inspired GNN solvers (dubbed PI-GCN and
PI-SAGE, respectively), and we assess the solution quality with the
normalized error $\epsilon = H_{\mathrm{Potts}}/|\mathcal{E}|$
quantifying the number of color clashes normalized by the number of
edges $|\mathcal{E}|$.  Accordingly, the quantity $\Xi = 1-\epsilon$ can
be regarded as the coloring accuracy achieved (that is the number of
edges without coloring conflicts divided by the total number of edges).
In addition we report upper bounds on the chromatic number, denoted as $\bar{\chi}_{\mathrm{GNN}}$. To this end we have implemented a simple search as well as randomized post-processing heuristic. 
For a given GNN solution, the latter tries to remove remaining color clashes at the expense of one additional color, by randomly going through existing clashes, and randomly assigning the new color to one of the two nodes at hand.
This process is repeated till a feasible coloring (with zero cost) has been found.
Our method has been implemented in python, leveraging the open-source
libraries Deep Graph Library \citep{dgl:19} and PyTorch (for GNN
handling), and NetworkX (for graph handling). All reported experiments
have been run on \texttt{p3.2xlarge} AWS instances, with 1 GPU, 8 virtual
CPUs, 81 GiB memory, 16 GiB GPU memory, with 2.3 GHz (base) and 2.7 GHz
(turbo) Intel Xeon E5-2686 v4 processors.

\textbf{COLOR graphs.} We study several benchmark instances
from the COLOR data set \cite{trick:02} 
which can be categorized as follows \citep{trick:02}:
\begin{itemize}

\item[(i)]{Book graphs: For a given work of literature, a graph is
created with each node representing a character. Two nodes are connected
by an edge if the corresponding characters encounter each other in the
book.  This type of graph is publicly available for Tolstoy's Anna
Karenina (anna), and Hugo's Les Mis\'erables (jean), among others.}

\item[(ii)]{Myciel graphs: This family of graphs is based on the
Mycielski transformation.  The Myciel graphs are known to be difficult
to solve because they are triangle free (clique number 2) but the
coloring number increases in problem size \citep{trick:02}.}

\item[(iii)]{Queens graphs: This family of graphs is constructed as
follows.  Given an $n$ by $n$ chessboard, a queens graph is a graph made
of $n^2$ nodes, each corresponding to a square of the board. Two nodes
are then connected by an edge if the corresponding squares are in the
same row, column, or diagonal.  In other words, two nodes are adjacent
if and only if queens placed on these two nodes can attack each other in
a single move. In all cases, the maximum clique in the graph is no more
than $n$, and the coloring value is lower-bounded by $n$.}

\end{itemize}
Our numerical results are summarized in Tab.~\ref{tab:color-graphs}.  We
consistently find sub-one-percent normalized errors ({\em i.e.},
$\epsilon <1\%$) across all COLOR instances, some of which have been
deemed as hard \citep{gualandi:11, kirovski:98}, with the
GraphSAGE-based architecture typically outperforming the GCN-based
baseline architecture. This observation appears to be in agreement with
existing literature \citep{alon:21} showing that GCN architectures tend
to be more susceptible to over-squashing (bottleneck) effects than other
GNN architectures. In this work, with its inherent neighborhood sampling
strategy, GraphSAGE is seen be more robust to potential over-squashing
effects as relevant for the larger and dense COLOR instances.  This
increased performance of PI-SAGE comes at a price of extended training
times, with per-epoch training times being $\sim 5$--$50$x longer than
for PI-GCN on the same graph. Whereas the PI-GCN model takes anywhere
from $\sim 0.167$ to $2$hr to train, the PI-SAGE model takes anywhere
from $\sim 1$ to $8$hr to train for the COLOR graphs considered. With
potentially multiple factors contributing to this disparity, a more
detailed analysis of this observation is left for future research.
Finally, we find that PI-SAGE performs on par with Tabucol
across the COLOR instances. 
In addition, the estimated chromatic numbers found with PI-SAGE are on par or better than the greedy baseline results. 
For example, for queen7-7 we find $\bar{\chi}_{\mathrm{GNN}}=7$, while $\bar{\chi}_{\mathrm{greedy}}=9$.
For solutions with non-zero cost we find that a simple post-processing heuristic can provide a fully purified (feasible) solution at the expense of a small number of colors, thereby providing a simple estimate for the chromatic number.
For example, solutions with just one remaining color clash, as is the case for queen8-8 and queen9-9, are trivial to purify at the expense of one color (yielding estimates of $\bar{\chi}_{\mathrm{GNN}}=q+1$).
For a larger number of color clashes, as is the case for queen11-11 and queen13-13, several iterations of this simple post-processing routine may be necessary till a feasible coloring is found. While further improvements may be possible through additional GNN runs at colors $q+1, q+2, \dots$, we observe on-par or better performance compared to the greedy baseline already with this simple post-processing only. 
The core of this post-processing routine is illustrated in Fig.~\ref{fig:postprocessing}.

\textbf{Citation graphs.} Next we provide results for
publicly-available, real-world citation graphs, with up to $n \sim 2
\times 10^{4}$ nodes. While Cora and Citeseer refer to networks of
computer science publications (with nodes representing publications and
edges referring to citations), the Pubmed citation network is a set of
articles related to diabetes from the PubMed database \citep{cora:00,
citeseer:08, pubmed:12}. Following Ref.~\citep{li:21}, the number of
available colors $q$ for the Cora, Citeseer, and Pubmed graphs has been
set to $5$, $6$ and $8$, respectively.  The results of our analysis are
displayed in Tab.~\ref{tab:citation-graphs}, 
with the greedy coloring algorithm providing optimal baseline results for these \textit{sparse} instances [with graph densities in the range $\sim (0.02 - 0.15)\%$]. 
We find that the basic
PI-GCN solver displays consistent, small errors $\epsilon \sim 10^{-4}$, close to
the global optimum. 
For the Cora and Citeseer graphs PI-GCN finds solutions with just one single color clash within $\sim 5 \times 10^3$ edges, while PI-SAGE finds optimal solutions at zero cost.
Note that local optimality has been verified for these solutions through a
series of simple local spin flips.  
Similarly, even for the largest
instance (Pubmed with $n \sim 2 \times 10^{4}$ nodes) we obtain a small
error of $\epsilon \sim 3 \times 10^{-4}$, while Tabucol fails to color
the graph within a 24 hours time limit \citep{li:21}. 
Conversely, we
find that PI-GCN converges for the Cora and Citeseer instances in $\sim
5$ to $40$min, respectively, while Pubmed takes $\sim 6.7$hr for
training completion. The comparatively long training time for Pubmed is
arguably due to separate logic being used to calculate the loss
function: Because of memory constraints on the training instances, here
we implemented sparse tensor calculations to ensure we would avoid
memory overload, at the expense of training time.  A more thorough
analysis together with the investigation of warm-starting (transfer
learning) strategies is left for future research.

Overall, we find that the general-purpose PI-GNN solver shows the potential to provide competitive coloring results compared to established, state-of-the-art heuristics such as the Tabucol algorithm \citep{hertz:87} or greedy coloring algorithms; in particular for large graphs (where Tabucol runtimes become extensively long) or dense graphs (where greedy algorithms may show performance drops).  
However, with the possibility to scale to problems with millions of nodes \citep{schuetz:21}, 
as well as the ability to solve other multi-class problems such as community detection or data clustering within the very same framework.

\section{Conclusion and Outlook}
\label{Conclusion}

In summary, we have shown how graph neural networks can be used to solve
graph coloring problems using insights from statistical physics as our
guiding principle. In our approach we frame graph coloring as a
multi-class (multi-color) node classification problem, with the Potts
model providing a canonical choice for the loss function with which we
train the GNN. Natively extending our framework as presented in
Ref.~\citep{schuetz:21}, we apply a relaxation strategy to the Potts
model by dropping integrality constraints on the decision variables in
order to generate a differentiable loss function with which we perform
unsupervised training on the node representations of the GNN. The GNN is
then trained to generate soft assignments to predict the likelihood of
belonging in one of $q$ classes, for each vertex in the graph. Post
training we use simple projection heuristics to find a coloring solution
consistent with the original problem.

Finally, we highlight possible extensions of research going beyond our
present work.  First, using the unifying framework established by the
Potts model, it would be interesting to apply our approach to other
multi-class problems such as community detection, data clustering, and
the minimum clique cover problem. Beyond that, one could apply our
approach to other large-spin (non binary) problems such as, for example,
variations of the Blume-Capel model which can be seen as a  generalization 
of QUBO to large-spin (i.e., multi-color) variables.  
Furthermore, one could systematically study the
potential existence of sharp coloring thresholds and the onset of
hardness with associated critical graph connectivities, for both
families of random graphs as done in Ref.~\citep{zdeborova:07} but also
more structured real-world problems.  Finally, there are several ways to
potentially boost the performance of our GNN-based optimizer.  For
example, one could explore alternative GNN implementations, potentially
in combination with graph rewiring techniques, as recently proposed and
analyzed in Ref.~\citep{topping:21}, thereby decoupling the training
graph from the original problem graph and providing additional GNN
design choices. In addition, post training one could replace the simple
deterministic (argmax) projection scheme used here with more
sophisticated strategies such as local search routines that further
refine the mapping from soft (probabilistic) class assignments to hard
(integer) variables.

{\bf Code availability statement}: An end-to-end open source demo version of the code implementing our approach has been made publicly available at \url{https://github.com/amazon-research/gcp-with-gnns-example}.

\begin{acknowledgments}

We thank 
% Fernando Brandao, George Karypis, 
Michael Kastoryano, Eric Kessler, Tyler Mullenbach, Nicola Pancotti, Mauricio Resende, Shantu
Roy, and Grant Salton for fruitful discussions.  
% This paper has been brought to you by the words ``{\em canonical},'' ``{\em specifically},'' and a $q$-coloring number between $1$ and $n$.

\end{acknowledgments}

\bibliography{refs}

\newpage \onecolumngrid \newpage { \center \bf \large  Supplemental Material for: \\  Graph Coloring with Physics-Inspired Graph Neural Networks \vspace*{0.1cm}\\  \vspace*{0.0cm} } 
\begin{center} Martin J. A. Schuetz,$^{1,2,3}$ J. Kyle Brubaker,$^{2}$ Zhihuai Zhu,$^{2}$ and Helmut G. Katzgraber$^{1,2,3}$ \\ 
	\vspace*{0.15cm} 
	\small{\textit{$^{1}$Amazon Quantum Solutions Lab, Seattle, Washington 98170, USA}} \\
	\small{\textit{$^{2}$AWS Intelligent and Advanced Compute Technologies, Professional Services, Seattle, Washington 98170, USA}} \\
	\small{\textit{$^{3}$AWS Center for Quantum Computing, Pasadena, CA 91125, USA}} 
	\vspace*{0.25cm} 
\end{center}

\setcounter{section}{0}
\setcounter{equation}{0} 
\setcounter{figure}{0} 
\setcounter{page}{1} 
\makeatletter 
\renewcommand{\theequation}{S\arabic{equation}} 
\renewcommand{\thefigure}{S\arabic{figure}} 
\renewcommand{\bibnumfmt}[1]{[S#1]} 
\renewcommand{\citenumfont}[1]{S#1} 

\section{Hyperparameters for PI-GNN on Benchmark Instances} \label{hypers}

\begin{table*} [h]
	\centering
	\begin{tabular}{| c | c | c c c c c c c |} 
		\hline
		graph & colors $q$ & embedding $d_{0}$ & layers $K$ & hidden dims [$d_1$, $d_2$, ...] & learning rate $\beta$ & dropout & number epochs & patience \\ [0.5ex] 
		\hline\hline
		anna & 11 & 43 & 1 & [22] & 0.03507 & 0.3298 & 100000 & 500 \\ 
		jean & 10 & 50 & 1 & [62] & 0.01663 & 0.3185 & 100000 & 500 \\ 
		myciel5 & 6 & 16 & 1 & [18] & 0.01333 & 0.3964 & 100000 & 500 \\ 
		myciel6 & 7 & 8 & 1 & [22] & 0.01779 & 0.2225 & 100000 & 500 \\ 
		queen5-5 & 5 & 77 & 1 & [32] & 0.02988 & 0.3784 & 100000 & 500 \\ 
		queen6-6 & 7 & 20 & 1 & [12] & 0.05105 & 0.3425 & 100000 & 500 \\ 
		queen7-7 & 7 & 67 & 1 & [12] & 0.02175 & 0.2339 & 100000 & 500 \\ 
		queen8-8 & 9 & 32 & 1 & [10] & 0.02728 & 0.2878 & 100000 & 500 \\ 
		queen8-12 & 12 & 107 & 1 & [23] & 0.01730 & 0.1796 & 100000 & 500 \\ 
		queen9-9 & 10 & 109 & 1 & [16] & 0.02636 & 0.3257 & 100000 & 500 \\ 
		queen11-11 & 11 & 75 & 1 & [25] & 0.04600 & 0.2974 & 100000 & 500 \\ 
		queen13-13 & 13 & 112 & 1 & [199] & 0.14426 & 0.1571 & 100000 & 500 \\ 
		[1ex] 
		\hline
	\end{tabular}
	\caption{
		Hyperparameters for results on COLOR graphs. All hyperparameters refer to our PI-SAGE architecture. All runs used the AdamW optimizer with default values (other than learning rate, as per the table).
		\label{tab:color-hypers}}
\end{table*}

\begin{table*} [h]
	\centering
	\begin{tabular}{| c | c | c c c c c c c |} 
		\hline
		graph & colors $q$ & embedding $d_{0}$ & layers $K$ & hidden dims [$d_1$, $d_2$, ...] & learning rate $\beta$ & dropout & number epochs & patience \\ [0.5ex] 
		\hline\hline
		Cora & 5 & 2342 & 1 & [3496] & 0.00556 & 0.0148 & 100000 & 500 \\ 
		Citeseer & 6 & 5127 & 1 & [2472] & 0.00983 & 0.0161 & 100000 & 500 \\ 
		Pubmed & 8 & 5137 & 1 & [6082] & 0.02966 & 0.1715 & 100000 & 500 \\ 
		[1ex] 
		\hline
	\end{tabular}
	\caption{
		Hyperparameters for results on citations graphs. All hyperparameters refer to our PI-GCN architecture. All runs used the Adam optimizer with default values (other than learning rate, as per the table).
		\label{tab:citations-hypers}}
\end{table*}

In this section, we provide details for the specific model
configurations (hyperparameters) as used to solve the COLOR and citation
graph instances with our physics-inspired GNN solver (PI-GNN).  The
results achieved with these model configurations are displayed in
Tab.~\ref{tab:color-graphs} and Tab.~\ref{tab:citation-graphs}; the
corresponding hyperparameters are given in Tab.~\ref{tab:color-hypers}
and Tab.~\ref{tab:citations-hypers}, respectively.

\end{document}